\documentclass[a4paper,12pt]{article}
\usepackage[english]{babel}
\usepackage[utf8]{inputenc}
\usepackage{fancyhdr}
\usepackage{afterpage}
\usepackage{graphicx}
\usepackage{hyperref}
\hypersetup{
    colorlinks=true,
    linkcolor=blue,
    filecolor=magenta,      
    urlcolor=cyan,
}
\newcommand\blankpage{%
    \null
    \thispagestyle{empty}%
    \addtocounter{page}{-1}%
    \newpage}
\begin{document}

\clearpage

\vspace*{\fill}
\begin{center}
\begin{minipage}{1\textwidth}
\title{Neurorehab: An Interface for Rehabilitation}
\author{Atul Dhingra, Adeboye A. Adejare Jr, Adam Fendler, \\Roopeswar Kommalapati}
\date{}
%\affil{Rutgers University} 
\maketitle
\end{minipage}
\end{center}
\vfill % equivalent to \vspace{\fill}
\afterpage{\blankpage}
\clearpage    

\section*{Abstract}
About 1 in 10 of the world population is affected by a disability in some form, amongst whom only 3 in 10 perform the recommended reahab tasks assigned without intervention. We are working on developing an interface to motivate and effectively encourage people. In our work, we leverage the fact that repetitive exercises can help people with motor disabilities due to the robust plasticity of the motor cortex. We investigate the role of repetitive activities for neurorehabilitation with the help of a non-invasive brain computer interface, formulated using immersive game design with Kinect v2.0 and Unity 3D. We also introduce a game design paradigm for dynamic difficulty adjustment for the patients.
\newpage

\section{Introduction}
Everyday tasks such as moving dishes comes naturally for those with regular human movement coordination as their movement goals and hand coordination align with little interference. For the people affected by motor and neurological disorders, the hand movements may not coordinate with the desired outcome for the task.  Scenarios such as sudden hand tremors,  or increased movement time between each action may  indicate  an irregularity of the motor control of the person.  To improve coordination, these patients go to rehabilitation centers to practice proper techniques in areas such a grabbing and movement.  These methods usually rely on physical therapists doing the physical movements while the patients coordinates the movements.  These efforts are not effective, physically exhausting for physical rehabilitation patients, and may not fully improve the patients coordination to either pre-injury status or regular human function pending on if condition is born or acquired.  A recent movement to improve health  is to integrate it with serious games, in the hopes of returning and improving the state of the patient using the game.

There has been a push for serious game development in health. Skip Rizzo and Pair developed a game to treat Post Traumatic Stress Disorder in soldiers\cite{PTSDSeriousGame}  Graafland, Schraagen, Schijven discovered multiple serious games that aided in improving surgical skills for doctors \cite{SurigcalSeriousGame}.  These areas improved the uses in their respectively category, because of these improvements many new application areas opened.  One area that this push recently entered is rehabilitation.

A myriad of  serious games for rehabilitation have came throughout the years.  Burke's group developed a method of using the Nintendo Wii interface to create a serious game rehabilitating stroke patients \cite{WiiStroke} Sucar's group also developed their own platform that used multiple interfaces such as a pressure sensitive marker system and a grip glove controller to work on improving rehabilitation efforts.\cite{GTMarker} While these methods helped rehabilitate patients, a better approach could come from using a different approach that does not require markers to measure movement.

We want to improve the rehabilitation serious game efforts by developing a serious game that will quantify movement.  While visual qualification varies between each tasks, we lack in numerical quantification of said hampered movements juxtaposed to routine movements. Having a marker tracking system such as from Burke and Sucar propagates this numerical quantification by having an intermediary interface between the user and the game.   With these issues in mind, how do we numerically quantify these irregularities, especially in regards to rehabilitation of a patient?   Creating an interactive simulation that provides an analogous setting to a normal household task with a marker-less tracking system , we can begin to quantify said irregularities in the context of rehabilitation efforts. The paper discusses about prior works and motivations in section 2. We talk about the various methods available in section 3, and about the experimental details in section 4.  We briefly talk about the interface design and relevant material in the appendix.\\

 A sequence of hand gesture is one of the most inherent features of human-human interaction\cite{Dhingra} used to interact with objects in their environment. The brain coordinates the position of the arm in 3D-space in relation to the object of interest.  In order to interact with the object, the hand must perform a grabbing motion. This is achieved by the brain relaying a series of inputs for the hand movement via action potentials to move an object from one position to another.  \\
% * <aj.adejare@rutgers.edu> 2015-12-05T15:58:49.809Z:
%
% >  /////////////
%
% Marking off where I moved some paragraphs
%
% ^.
\section{Prior Works and Motivations}
Motor disabilities cause limitation in the fine motor skills, strength and range of motion\cite{YoungAdults}. This inhibits the diurnal activities of the patients and has a negative impact on the patients. It is seen that repetitive exercises can help people with motor disability\cite{YoungAdults}. This ensues the role of physiotherapists to assist the patients in these tasks.But due to lack of motivation only 31\% of the people perform the recommended exercises (Shaughnessy,Resnick,\&Macko,2006). One way to remedy this is to engage the patients using one-to-one interaction with the physiotherapists, but this option is not economical in general. This results in further atrophy of muscles and insufficient muscle endurance. To remedy such a situation, Kleim et al (‘03) suggest that occupational therapy can sufficiently simulate brain to remodel itself\cite{YoungAdults}. This allows the patients to have a better motor control. This motivated to identify motivating and effective methods for encouraging patients suffering from motor disabilities. \\

The rise of sensors and computing technologies for making the rehabilitation process more intuitive enabled to eliminate the reliance on trained Physiotherapists and the human error introduced along with. The two main families of sensors used for human motion capture for rehabilitation are are optoelectronics and nonoptoelectronics\cite{Review}. The first set contains the sensors that are marker-based system, i.e they require some of the parts to be affixed/controlled by the user in order to record the response for physiotherapy. These include inertial sensors, magnetic sensors, wearable sensors, and mechanical sensors. For instance Anguera et al\cite{Nature} used Joystick to interact with the 3-D custom-game, NeuroRacer. This kind of a framework can cause inconvenience to the user apart from the fact that it requires special infrastructure to set up such as designated areas of operations. The latter set consists primarily of methods that use marker-less system including Skin color based tracking, depth based tracking etc. It enables the users to interact using natural gestures, as in the case of Kinerehab\cite{YoungAdults}. This not only creates a more immersive environment for the patients, but it also helps to track the progress that can be gauged for accuracy by medical professionals. This motivated us to work towards our project. For now we are focusing our attention towards children, as only a few traditional physiotherapy techniques exist that are designed keeping in mind their  needs. We have developed a brain computer interface to aid neuro-rehabilation using Kinect and Unity 3D. This also provides an interface that can be used for assessing performance measure.

\section{Methods}
For our work, we used the Microsoft Kinect, a depth-sensing motion camera capture device which uses both a RGB camera as well as a laser to detect and process coordinates. The laser in the Kinect acquires the depth information of the user, using it to process a person’s forward and backwards movement in relation to the Kinect. Through the use of the Kinect, the game software can track a person’s movement and give us a model of a person’s actions when playing the game.\\

The Kinect was essential in allowing us to create a markerless, gamefied system for rehabilitation. There are many advantages afforded by markerless systems when compared to marker-based systems. They allow patients to feel a lower degree of social stigma, which may otherwise negatively affect rehabilitation. They are also cheaper in general, and can be re-purposed or constructed by modifying and reprogramming other technology. While there is a decrease in accuracy as the lack of markers allow a greater degree of error propagation to occur, for the purpose of upper body rehabilitation, this difference is negligible.\\

The Kinect offers a fairly low level of error propagation, and can track targets between 0.5 to 5 meters away. Its random error accuracy ranges from a few millimetres of error at 0.5m to 4cm of error at 5m. It also has a sampling frequency of approximately 30 frames per second. This is sufficient for the purpose of recording body movement for the game.\\

The Unity Game engine provides a suite of development tools for game creation.  For this game, we used multiple free libraries provided by Unity and other third parties in order to establish the basic environment settings. These included the Kinect Free Software Development Kit and the Raw Mocap Data for Mecanim library. We also utilized the MS-Kinect API and the Final Inverse-Kinematics library available for use in the Unity Asset Store to help establish the parameters and setup variables needed in the project.\\

The module of the Unity Game Engine used was Unity 5 3-D, the version of Unity which supports the creation of 3-dimensional games. The Kinect Free SDK was essential in providing the ability to integrate the Kinect hardware with the Unity Game Engine, and provided commands to interact with the raw Kinect data. We were able to import and map the Kinect inputs to an avatar via the Raw Mocap Data library, as well as create the animations to get the avatar moving. In addition, the MS-Kinect API provided gesture support, and was helpful in establishing the grabbing gesture necessary to move the target.\\

Finally, the Final Inverse-Kinematics library was one of the essential APIs used. It allowed the skeleton and joint structure of the avatar to be oriented towards the target ball, making it possible for the posture of the hand to dynamically change corresponding to the location of the ball. This increased the realism and accuracy of the model of the hand for the game.

\section{Experiment}
The objective of this experiment is to understand how the brain behaves in regard to a person in rehabilitation completing a task of grabbing and moving an object.  Furthermore, we want to discern the traits between someone rehabilitating from an impairment in comparison to a person currently not rehabilitating.  To find out these traits, we want to take an approach that can record the information of doing the task, while allowing us to observe with as little obstruction as possible.  Having these conditions will allow us to see how the brain achieves these goals in a relaxed state.  Looking at current technology, a method to achieve these goals is through sensors; one particular sensor that can show us these states is the Microsoft Kinect.\\

For this project, we will focus on a person’s right arm for tracking.  Culturally, the right hand is used for most interactions and daily tasks. The right hand is also, for a significant part of of the population, the dominant hand used for handwriting and other motor tasks.  Using the right hand, we can concentrate on getting the information for a small sample state while prototyping for the future.\\

For the environment, we wanted to emulate natural home settings for the player moving objects.  Thus, constructing a table gives the player a frame of reference as to how the object should naturally move in space and adds a sense of realism for the task. Using the table also provides a coordinate reference for the object in relationship to the scene, showing where an error would occur if the object dropped unexpectedly.\\

The object of the game requires users to move a ball to the target goal.  The ball will first need to be grabbed and then moved in space to the desired area.  From the desired area the ball would drop from the user’s command, having it hit the target.  The user would repeat this motion multiple times throughout the game until reaching the allocated repetition time.  This goal would then reduce in size and the person would go through the same motion again; however, should the user miss the goal repeatedly, the goal target would increase in size, allowing for an adaptive game difficulty.\\

Performing the task, the user would have two states once the ball is grabbed.  If the task is completed successfully, a user would get a visual cue of "Success" as well as an auditory cue to confirm the successful drop into the target area.  If the the task was unsuccessful, a visual cue of "Try Again" would emerge, along with an auditory cue that would play, attempting to coax the user to try again using the correct method to pick up and drop the ball in its correct place. \\ 

Reducing the goal size, we want to improve on the precision and accuracy of the player’s movement.  By seeing if they can drop the ball correctly on the target’s area, we can obtain the accuracy of the movement.  Observing the repetition for the movement and the general area in which the player places the ball we can measure the precision of the player’s movements. 

\section{Conclusions and Future Works}
We developed a rehabilitation simulation directed for children and young adults. We have therefore focused our attention to positive feedback for the system. This system can be easily extrapolated to rehabilitation of adults, which opens other avenues of feedback in form of haptics(negative feedback). In future we would like to inspect data from the system so as to provide a module for the patients to track their progress. 

\section{Acknowledgment}
We would like to thank our advisors Konstantinos Michmizos and Mubbasir Kapadia for the invaluable input during the project, Kostas Bekris (CBIM) for providing us with the Kinect v2.0 for the implementation of our interface for rehabilitation.  We thank Rumen Filkov and Hai Xuan Pham for technical support. 

\section{Appendix}
The game menu design has 3 options, namely to start the game, load the saved data, and delete the saved data. The game itself was designed around a table containing a ball, and a target circle. The avatar stands in front of the table, and the current score is shown in the top-right corner.
\subsection{Interface Design}
\begin{center}
\includegraphics[width=12cm,height=7cm,scale=0.35]{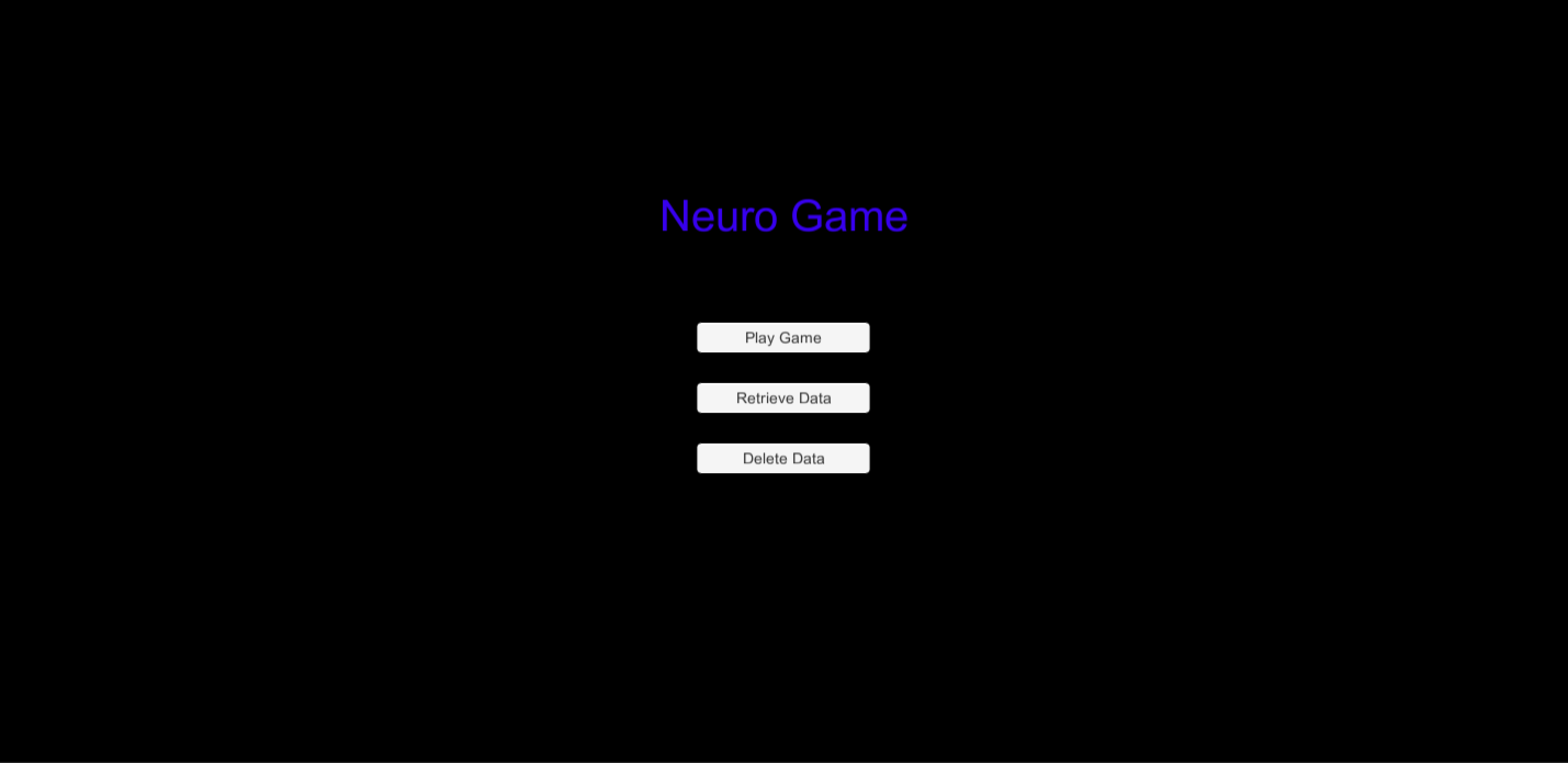}\\
\includegraphics[width=12cm,height=7cm,scale=0.35]{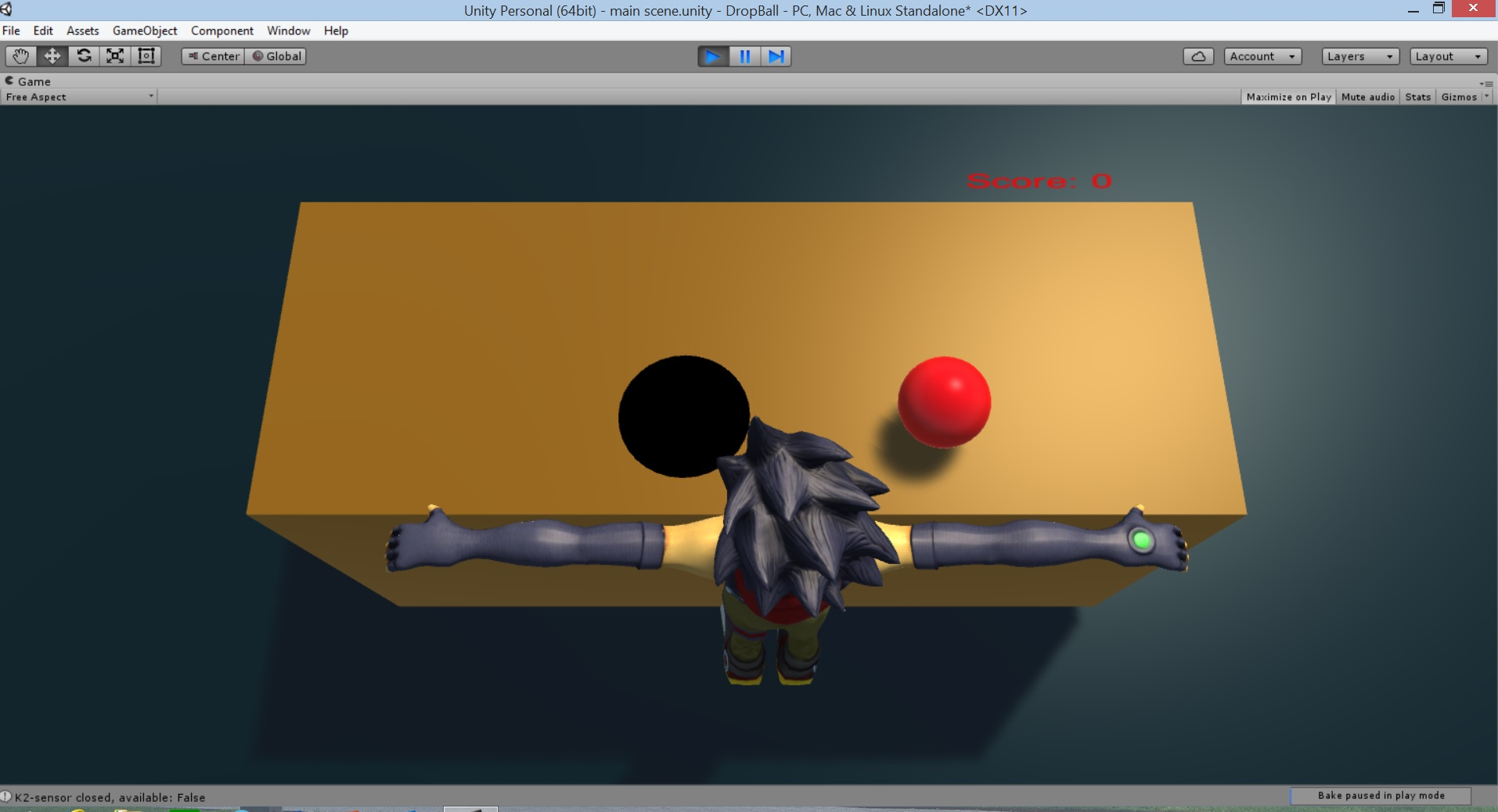}
\end{center}
\subsection{Code Implementation}
The project was implemented using C\# language. The code is available at \url{https://github.com/dhingratul/Neurorehabilitation-Gamification}

\end{document}